\documentclass[runningheads]{llncs}
\usepackage[T1]{fontenc}
\usepackage{nicematrix,booktabs,caption}
\usepackage{verbatim}
\usepackage{graphicx}
\usepackage{hyperref}
\usepackage{multirow, multicol}
\usepackage{booktabs}
\usepackage{amsmath}
\usepackage{float}
\usepackage{amsfonts}
\usepackage{bm}
\usepackage{marvosym}
\usepackage{ulem}
\usepackage{cite}
\usepackage{array}
\usepackage{xcolor}
\usepackage{arydshln}
\definecolor{babyblue}{rgb}{0.54, 0.81, 0.94}

\newcolumntype{L}[1]{>{\raggedright\let\newline\\\arraybackslash\hspace{0pt}}m{#1}}
\newcolumntype{C}[1]{>{\centering\let\newline\\\arraybackslash\hspace{0pt}}m{#1}}
\newcolumntype{R}[1]{>{\raggedleft\let\newline\\\arraybackslash\hspace{0pt}}m{#1}}

\begin{document}

\newcommand{\key}[1]{{\textcolor{blue}{\textbf{[#1]}}}}

\title{3D Consistency Optimization for Self-Supervised Monocular Video Depth Estimation}

\author{
Yuanye Liu\inst{1}\protect\footnotemark[2] \and
Ke Zhang\inst{2}\protect\footnotemark[2] \and
Junzhe Jiang\inst{1} \and
Li Zhang\inst{1} \and
Vishal Patel\inst{2} \and
Xiahai Zhuang\inst{1}\protect\footnotemark[5]
}

\authorrunning{Y. Liu et al.}

\institute{
Fudan University, Shanghai, China \and
Johns Hopkins University, Baltimore, USA
}

\authorrunning{Yuanye Liu et al.}
\makeatletter
\renewcommand\@fnsymbol[1]{%
  \ensuremath{%
    \ifcase#1\or
      *\or
      \dagger\or
      \ddagger\or
      \S\or
      \text{\Letter}\or
      \mathparagraph\or
      \|\or
      **\or
      \dagger\dagger
    \fi
  }%
}
\makeatother
\renewcommand{\thefootnote}{\fnsymbol{footnote}}
\footnotetext[2]{These two authors contributed equally.}
\footnotetext[5]{Corresponding authors: Xiahai Zhuang (zxh@fudan.edu.cn)}

\maketitle
\begin{abstract}
Reliable monocular video depth estimation is crucial for downstream 3D reasoning and embodied AI in endoscopic navigation. However, existing self-supervised approaches typically treat video frames independently or rely on weak temporal regularization. These methods, lacking a holistic perception of the underlying 3D scene, inevitably suffer from geometrically inconsistent predictions and severe cross-frame drift. To address these limitations, we introduce a new paradigm that recasts sequential video depth estimation as an unconstrained multi-view 3D reconstruction problem, enabling full exploitation of the powerful geometric priors embedded in recent 3D foundation models. The core of our approach is a 3D consistency optimization framework driven by three constraints: image-level photometric rendering, explicit world-coordinate geometric alignment, and multi-scale temporal gradient consistency. Such unified optimization elegantly anchors isolated frames to a globally coherent 3D structure. Our method has been validated in both the self-supervised training scenarios and challenging zero-shot clinical environments. Results show that the proposed approach achieves state-of-the-art spatial accuracy, outperforming the frame-based, video-based depth estimators and the multi-view 3D reconstruction baselines.
\end{abstract}

\section{Introduction}
Accurate dense depth estimation for endoscopic video is a fundamental enabler for downstream 3D reasoning in minimally invasive surgery, paving the way for advanced spatial computing and embodied surgical AI, including dense reconstruction \cite{huang2024endo4dgs,wang2024endogslam,yang2023neurallerplane}, navigation \cite{shao2022selfsupervisedendoscopy,wei2024enhancedscaleaware,C_2025MICCAI_navigation}, and robot-assisted guidance and surgical scene understanding \cite{C_2024MICCAI_endodac,C_2025MICCAI_endodav}. 
However, reliable clinical deployment remains severely hindered by hardware and physical constraints.
Most standard clinical endoscopes are monocular and inherently lack reliable camera pose tracking,
making pose-free monocular depth estimation an unavoidable challenge 
in real-world operating rooms~\cite{C_2017CVPR_pose-free-1,C_2025CVPR_pose-free-2}.

Existing self-supervised approaches have attempted to address this challenge by learning depth from monocular sequences~\cite{C_2024MICCAI_endodac,C_2025MICCAI_endodav}. 
However, they predominantly treat video sequences as a collection of independent frames or apply weak temporal regularization.
These local constraints fail to capture the holistic 3D geometry of the underlying surgical scene. 
Consequently, in the challenging endoscopic conditions, small frame-wise prediction errors rapidly accumulate, leading to severe 3D geometric drift and temporal flickering, heavily undermining the reliability of any downstream 3D reconstructions~\cite{C_2025MICCAI_Endo3R}.

To overcome these fundamental limitations, we propose a paradigm shift in how monocular endoscopic videos are processed. 
Inspired by the remarkable geometric capabilities of recent multi-view foundation models~\cite{wang2025vggt,jiang2025anysplat,C_2025MICCAI_foundation}, we propose viewing an endoscopic video not merely as a time-bound sequence, but as a collection of unconstrained multi-view observations anchored to a shared 3D scene
By treating isolated video frames as multi-view samples, we can fully leverage the strong spatial priors of modern 3D foundation models. 
This perspective elegantly transforms the ill-posed monocular video depth estimation task into a more robust multi-view 3D reconstruction problem, effectively bridging the gap between foundation models and pose-free clinical endoscopy.

Building upon this insight, we present a novel 3D consistency optimization framework for self-supervised monocular video depth estimation,
which explicitly enforces cross-frame geometric coherence without requiring ground-truth camera poses. 
Our framework jointly optimizes dense depth maps and camera poses by anchoring them to a unified world coordinate system. This is achieved through a differentiable 3D Gaussian splatting~\cite{J_2023ACMToG_3DGS,C_2025MICCAI_3dgs,C_2025MICCAI_SurgTPGS} mechanism driven by three synergistic constraints: 
(1) an image-level photometric rendering loss that ensures appearance consistency; 
(2) an explicit world-coordinate geometric alignment loss that forces corresponding 3D points from different frames to project consistently, acting as the primary mechanism to eliminate spatial drift; 
and (3) a multi-scale temporal gradient consistency loss that preserves geometry-aware discontinuities and suppresses high-frequency flickering.

We rigorously evaluate our framework against representative baselines under both standard self-supervised settings and challenging zero-shot clinical scenarios.
Our contributions are summarized as follows:
\begin{enumerate}
\item We introduce a novel self-supervised framework that reformulates monocular video depth estimation as a pose-free, multi-view 3D consistency optimization problem, unlocking the potential of 3D foundation models for clinical endoscopy.

\item We propose a unified optimization strategy featuring explicit world-coordinate alignment and multi-scale temporal gradient constraints, effectively anchoring sequential frames to a globally coherent geometry to eliminate cross-frame 3D drift.

\item We achieve state-of-the-art accuracy on public benchmarks, demonstrating robust zero-shot generalization to unconstrained clinical environments.
\vspace{-1em}
\end{enumerate}

\begin{figure}[!t]
    \centering
    \includegraphics[width=\textwidth]{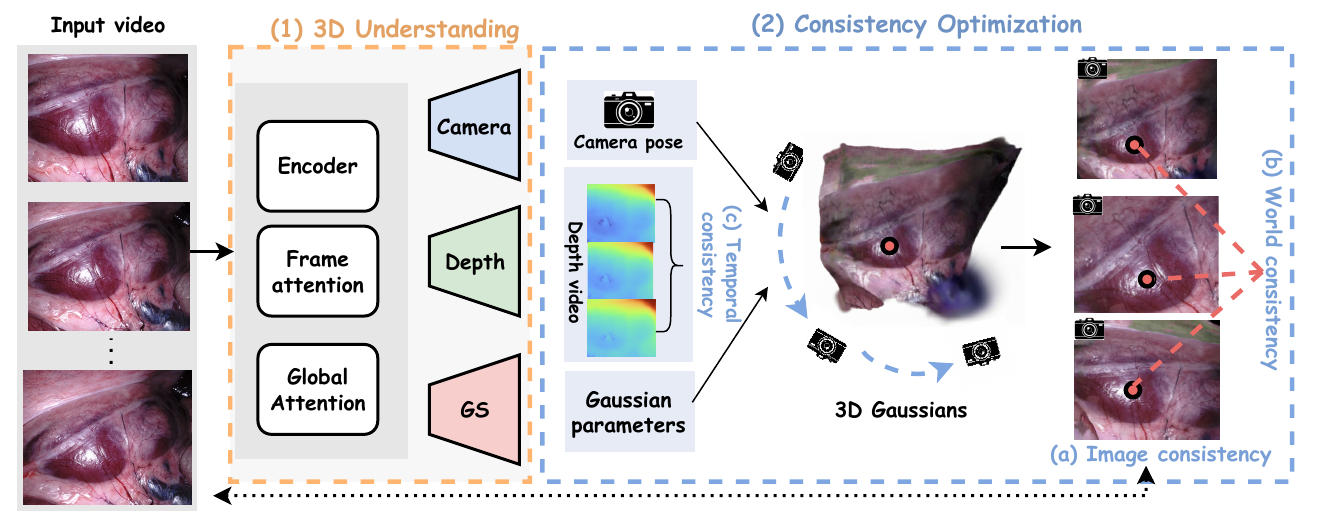}
    \caption{Overview of our framework. We first perform 3D understanding using a geometry transformer, followed by consistency optimization, which includes (a) image-level consistency for RGB rendering supervision, (b) world consistency for 3D coordinate alignment, and (c) temporal consistency to suppress flickering.}
    \label{fig:pipeline}
    \vspace{-0.5em}
\end{figure}
\section{Methodology}\label{sec:method}

\subsection{Problem Formulation and Scene Representation}
Given a monocular endoscopic video sequence $\mathcal{V} = \{ I_t \}_{t=1}^T$ where $I_t \in \mathbb{R}^{H \times W \times 3}$, our goal is to estimate dense depth maps in a fully self-supervised manner.
Instead of treating the video as isolated frames, we formulate this task as a multi-view 3D scene reconstruction problem. 

We represent the underlying surgical scene using a set of 3D Gaussian primitives $\mathcal{P} = \{p_i\}_{i=1}^M$. Each primitive $p_i$ is parameterized by its 3D center, opacity, anisotropic covariance (scale and rotation), and spherical harmonic coefficients for view-dependent appearance. 
Concurrently, our framework estimates the camera intrinsic matrix $K$ and the extrinsic poses $T_t \in SE(3)$ for each frame. To resolve gauge ambiguity and unify the global coordinate system, we anchor the 3D space to the first frame by setting $T_1 = \mathbf{I}$.

Given the 3D Gaussians $\mathcal{P}$ and the predicted camera parameters $\{T_t, K\}$, we employ differentiable Gaussian splatting to render both the expected RGB image $\hat{I}_t$ and the rendered depth map $\hat{D}_t$ at frame $t$:
\begin{equation}\label{eq:render}
    \hat{I}_t, \hat{D}_t = \mathcal{R}(\mathcal{P}, T_t, K),
\end{equation}
where $\mathcal{R}(\cdot)$ denotes the differentiable rendering operator. 
These rendered outputs serve as our final predictions and form the basis for our self-supervised consistency optimization.

\subsection{3D Geometry Understanding Network}
To extract global 3D geometry from the sequential input, our architecture builds upon recent multi-view transformer designs~\cite{jiang2025anysplat,wang2025vggt} as illustrated in Fig.~\ref{fig:pipeline}. 
Input images are first tokenized using DINOv2~\cite{oquab2023dinov2} features. These tokens are processed by an Alternating-Attention Transformer, which interleaves frame-level intra-view attention with global inter-view attention to enforce spatial-temporal coherence. 

The network outputs are generated via three specialized heads:
(1) The \textbf{Camera Head} refines a learnable camera token to predict the per-frame extrinsic poses $T_t$ and intrinsics $K$. 
(2) The \textbf{Depth Head} predicts an initial per-pixel depth map $D_t^{\text{init}}$. Crucially, this intermediate depth is exclusively utilized to back-project image pixels into 3D space, providing the necessary spatial initialization for the Gaussian centers. 
(3) The \textbf{Gaussian Head} fuses transformer-derived features with shallow CNN appearance features to predict the remaining attributes (opacity, orientation, and spherical harmonics) for each back-projected primitive. Together, these modules map the 2D video into a unified, differentiable 3D Gaussian representation.

\subsection{3D Consistency Optimization Framework}
Since ground-truth depth is unavailable, we optimize the network by imposing consistency constraints directly on the rendered outputs $\hat{I}_t$ and $\hat{D}_t$. By jointly optimizing image space, 3D space, and temporal transitions, we effectively regularize the entire 3D representation.
Crucially, this consistency optimization is strictly a training-time objective. 
During inference, our model predicts depth in a single feed-forward pass without any per-scene test-time optimization (TTO).

\vspace{0.5em}
\noindent\textbf{Image-level consistency.} 
To ensure the reconstructed 3D Gaussians accurately capture the appearance and structure of the scene, we enforce a photometric reconstruction loss between the rendered image $\hat{I}_t$ and input frame $I_t$:
\begin{equation}
    \mathcal{L}_{\text{rgb}} = \sum_{t=1}^T \| I_t - \hat{I}_t \|_1.
\end{equation}
Minimizing this image-level error implicitly forces the model to learn geometrically plausible 3D structures and appearances.
\vspace{0.5em}
\begin{figure}[!t]
    \centering
    \includegraphics[width=\textwidth]{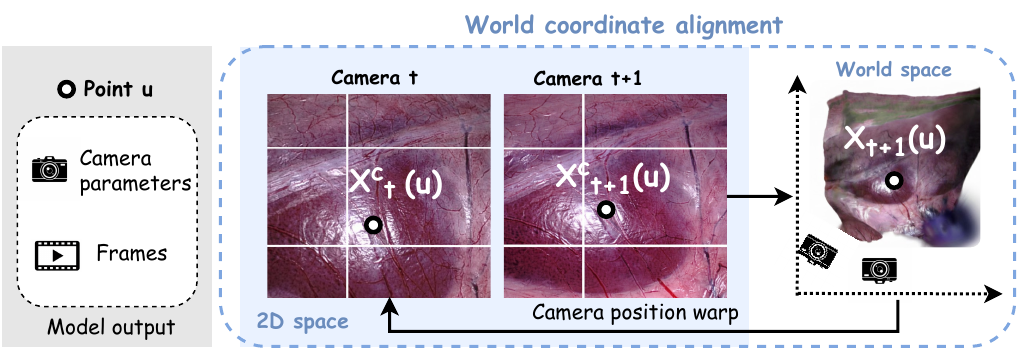}
    \caption{Illustration of 3D world coordinate consistency loss.}
    \label{fig:illustration}
\end{figure}

\noindent\textbf{World-Coordinate Geometric Consistency.} 
As shown in Fig.~\ref{fig:illustration}, to explicitly suppress spatial drift, we enforce strict multi-view geometric coherence on the rendered depth maps. For a pixel $u$ at frame $t$, we unproject its rendered depth $\hat{D}_t(u)$ into a 3D point in the world frame:
\begin{equation}
    X_t(u) = T_t \left( \hat{D}_t(u) K^{-1} \tilde{u} \right).
\end{equation}
We then project this global point $X_t(u)$ onto a neighboring view $t'$ with stride $\Delta$ to obtain its 2D coordinate $u_{t \rightarrow t'} = \pi\big(T_{t'}^{-1} X_t(u), K\big)$, where $\pi(\cdot)$ denotes the camera projection. 
To robustly handle occlusions and out-of-bounds projections, we compute the consistency loss exclusively over a valid pixel set $\Omega_{t,t'}$. A projected point is valid only if it falls within the target image boundaries and has a strictly positive depth. 
Furthermore, to mitigate the impact of residual dynamic tissue motions and unmasked depth outliers, we apply a robust Huber penalty to the depth discrepancy. The 3D world consistency loss is formulated as:
\begin{equation}
    \mathcal{L}_{\text{3D}} = \sum_{t,t'} \frac{1}{|\Omega_{t,t'}|} \sum_{u \in \Omega_{t,t'}} \rho \Big( \hat{D}_{t'}(u_{t \rightarrow t'}) - z(X_t(u)) \Big),
\end{equation}
where $z(X_t(u))$ is the depth of $X_t(u)$ along the ray of camera $t'$, and $\rho(\cdot)$ denotes the Huber loss function. Here, $\Omega_{t,t'}$ defines the set of valid pixels, and $|\Omega_{t,t'}|$ is its cardinality.

\vspace{0.5em}
\noindent\textbf{Temporal Consistency.} 
Video sequences naturally exhibit temporal continuity, with scene geometry typically undergoing smooth transitions over short intervals. 
To exploit this property and suppress temporal flickering, we introduce a multi-scale temporal gradient consistency loss that encourages adjacent frames to maintain similar depth-edge structures.
Given a predicted depth map $D_t \in \mathbb{R}^{H \times W}$ at time $t$, we measure the spatial gradient discrepancy between $D_t$ and a neighboring frame $D_{t+\Delta}$ exclusively within a valid-pixel mask. 
Specifically, at each scale $s$, we downsample the depth maps by a factor of $2^s$ and compute the first-order finite differences along the $x$ and $y$ spatial dimensions. 
We then penalize the absolute mismatch between these gradients. Averaged across all temporal pairs, the loss is formulated as:
\begin{equation}
    \mathcal{L}_{\text{temp}} = \frac{1}{N} \sum_{t=1}^{N} \sum_{s=0}^{S-1} \left( \| \nabla_x D_t^{(s)} - \nabla_x D_{t+\Delta}^{(s)} \|_1 + \| \nabla_y D_t^{(s)} - \nabla_y D_{t+\Delta}^{(s)} \|_1 \right) \odot \Omega_t,
\end{equation}
where $\Omega_t$ denotes the valid-pixel mask for the frame pair at time $t$. This term effectively regularizes temporal flicker while preserving geometry-aware discontinuities, yielding highly stable depth predictions across frames.

The final training objective is a weighted combination of all loss terms:
\begin{equation}
\mathcal{L} =
\lambda_{\text{1}} \mathcal{L}_{\text{rgb}} +
\lambda_{\text{2}} \mathcal{L}_{\text{3D}} +
\lambda_{\text{3}} \mathcal{L}_{\text{temp}}
\end{equation}

By jointly optimizing these complementary constraints, our method integrates geometric reasoning in 3D space, image-level consistency, global alignment in the 3D world coordinate system, and temporal coherence across frames, enabling more robust self-supervised video depth estimation.

\section{Experiments}
\subsection{Setup}

\noindent\textbf{SCARED.} Captured via a da Vinci Xi surgical system, this dataset contains 35 high-resolution endoscopic videos (22,950 frames) with projector-measured ground-truth depth, camera poses, and intrinsics.
Following~\cite{C_2025MICCAI_endodav}, we split the sequences into 24 for training, 3 for validation, and 8 for testing to rigorously evaluate spatial accuracy and temporal consistency.

\noindent\textbf{Hamlyn.} Featuring diverse real-world \textit{in vivo} endoscopic and laparoscopic procedures, Hamlyn captures complex clinical challenges like dynamic tissue motion and irregular camera trajectories. We utilize the rectified version from~\cite{J_2021_hamlyn}, which provides ground-truth depth maps generated via Libelas. All sequences from this dataset are exclusively used for zero-shot evaluation.

\noindent\textbf{Evaluation metrics.} 
Following prior work on depth estimation~\cite{C_2024MICCAI_endodac,C_2025MICCAI_endodav}, we report five commonly used error and accuracy metrics, including Abs Rel, Sq Rel, RMSE, RMSE log, and $\delta_1$. 
To fairly evaluate affine-invariant depth predictions, we align the predicted depth with the ground truth using a median scaling following~\cite{C_2025MICCAI_endodav}. Finally, we compute the metrics for each scene and report the average results across all scenes.

\noindent\textbf{Implementation details.}
Our method is implemented in PyTorch~\cite{paszke2019pytorch} and trained on a single NVIDIA A100 GPU.
We use an Alternating-Attention Transformer with 24 layers. 
The weights are initialized from AnySplat~\cite{jiang2025anysplat}. 
During fine-tuning, the depth and Gaussian heads are fully trainable. 
To ensure stable initialization, the camera head is kept frozen during the initial warmup phase and is subsequently unfrozen for joint optimization, while all other pre-trained components remain frozen.
We optimize with AdamW~\cite{loshchilov2017adamw} for 20000 iterations using a cosine learning-rate schedule with 2000 warmup iterations. The peak learning rate is $2\times10^{-4}$, and we apply a $0.1\times$ learning-rate multiplier to parameters. Each iteration samples an endoscopic video clip of 16 frames. 
For the consistency loss, the temporal stride $\Delta$ is set to 6, and the loss weights are balanced as $\lambda_1=\lambda_2=\lambda_3=1$.

\begin{table*}[t]
\centering
\caption{Quantitative depth estimation results on SCARED and Hamlyn datasets.
$\downarrow$ indicates lower is better and $\uparrow$ indicates higher is better.}
\label{tab:comparison1}
\resizebox{\textwidth}{!}{
\begin{NiceTabular}{llcccccc}
\CodeBefore
\rowcolor{babyblue!20}{6}
\rowcolor{babyblue!20}{11}
\Body
\toprule
Dataset& Method & Year 
& Abs Rel $\downarrow$ 
& Sq Rel $\downarrow$ 
& RMSE $\downarrow$ 
& RMSE log $\downarrow$ 
& $\boldsymbol{\delta}_1$ $\uparrow$ \\
\midrule
\multirow{5}{*}{SCARED}& EndoDAC~\cite{C_2024MICCAI_endodac} 
& 2024 
& 0.201 
& 5.163 
& 16.421 
& 0.238 
& 0.653 
\\

&VDA~\cite{C_2025CVPR_VDA} 
& 2025 
& 0.241 
& 7.702 
& 18.673 
& 0.287 
& 0.597 
\\

&{EndoDAV~\cite{C_2025MICCAI_endodav}} 
& 2025
& {0.156} 
& {3.113} 
& {12.257} 
& {0.182} 
& {0.761} 
\\

&{AnySplat~\cite{jiang2025anysplat}}
& 2025 
& {0.100} 
& {2.227} 
& {10.400} 
& {0.255} 
& {0.902} 
\\

&{Ours} 
& -- 
& \textbf{0.060} 
& \textbf{0.471} 
& \textbf{5.503} 
& \textbf{0.079} 
& \textbf{0.978} 
\\

\midrule
\multirow{5}{*}{Hamlyn}&EndoDAC~\cite{C_2024MICCAI_endodac} 
& 2024 
& 0.240 
& 6.998 
& 17.240 
& 0.304 
& 0.589 
\\

&VDA~\cite{C_2025CVPR_VDA} 
& 2025 
& 0.389 
& 19.308 
& 23.005 
& 0.333 
& 0.513 
\\

&{EndoDAV~\cite{C_2025MICCAI_endodav}} 
& 2025 
& {0.212} 
& {5.040} 
& {16.759} 
& {0.276} 
& {0.595} 
\\

&{AnySplat~\cite{jiang2025anysplat}}
& 2025 
& {0.299} 
& {17.133} 
& {29.938} 
& {0.646} 
& {0.588} 
\\

&{Ours} 
& -- 
& \textbf{0.108} 
& \textbf{1.554} 
& \textbf{10.535} 
& \textbf{0.136} 
& \textbf{0.883} 

\\

\bottomrule
\end{NiceTabular}
}
\vspace{-1.5em}
\end{table*}

\begin{figure}[!t]
    \centering
    \includegraphics[width=\textwidth]{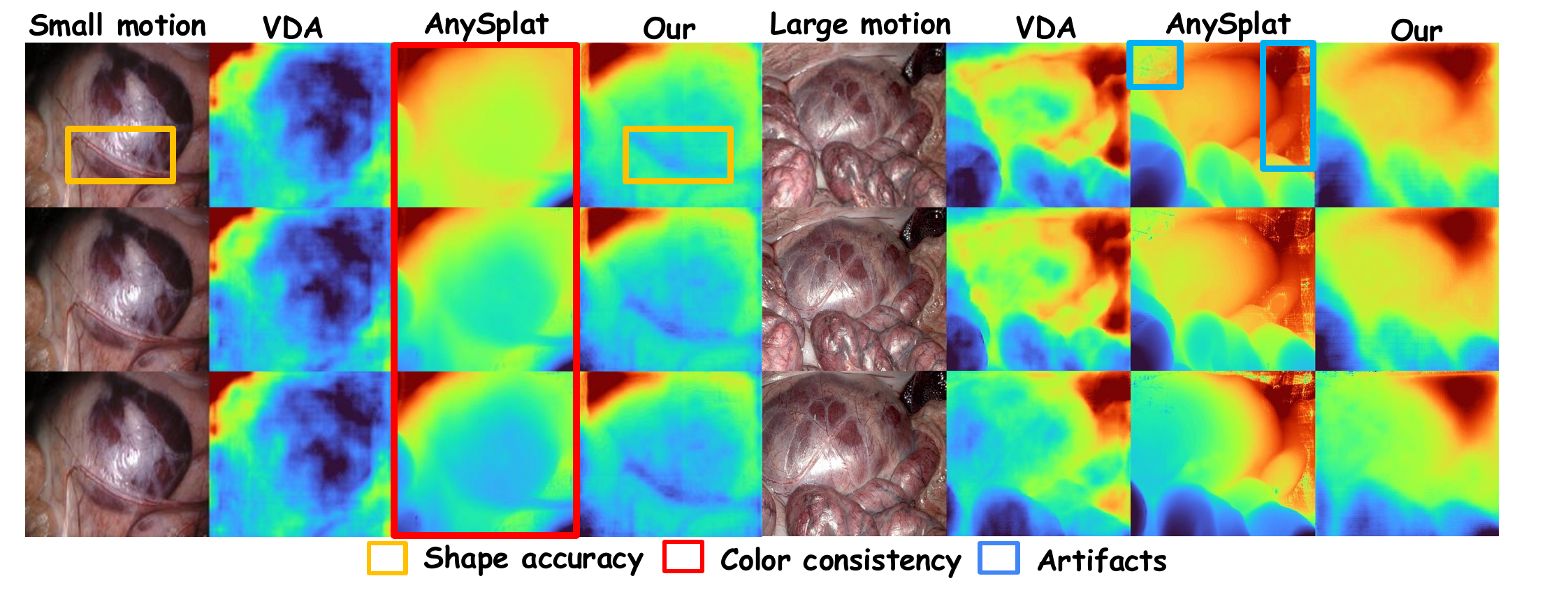}
    \caption{Qualitative comparison of our method with existing depth estimation approaches on two typical cases.}
    \label{fig:visualization}
    \vspace{-1em}
\end{figure}

\subsection{Results}
We compare our method with several representative depth estimation approaches on two public surgical datasets, SCARED and Hamlyn, including recent methods such as EndoDAC~\cite{C_2024MICCAI_endodac},
VDA~\cite{C_2025CVPR_VDA},
EndoDAV~\cite{C_2025MICCAI_endodav}
and AnySplat~\cite{jiang2025anysplat},  
which share our specific task setting under identical data splits.
Quantitative comparisons are reported in Table~\ref{tab:comparison1}. 
Overall, our approach consistently outperforms prior methods on both datasets, demonstrating strong generalization in challenging endoscopic scenarios.
On SCARED, our method achieves the best performance across all evaluation metrics. In particular, it significantly reduces the error metrics (Abs Rel, Sq Rel, RMSE, and RMSE log) while achieving the highest accuracy in $\delta$. Compared with the strongest previous method, AnySplat, our approach improves Abs Rel from 0.100 to 0.060 and RMSE from 10.400 to 5.503, indicating more accurate and stable depth predictions. These improvements highlight the effectiveness of the proposed consistency optimization framework.
On Hamlyn, our method also achieves the best results across all metrics, showing strong generalization in zeroshot evaluation.
For example, Abs Rel decreases from 0.299 (AnySplat) to 0.108, and RMSE is reduced from 29.938 to 10.535, while $\delta$ increases to 0.883.

As visualized in Fig.~\ref{fig:visualization}, our method better preserves scene structure and avoids abnormal errors and temporal inconsistencies, as highlighted by the red boxes.
This suggests that our method generalizes well to different surgical environments and scene characteristics.
Overall, both quantitative and qualitative comparisons demonstrate that our approach provides more accurate and reliable depth estimation compared with existing methods.

\begin{table*}[t]
\centering
\caption{Ablation study of proposed components on the SCARED dataset.}
\label{tab:ablation_1}
\resizebox{\textwidth}{!}{
\begin{NiceTabular}{ccccccccc}
\CodeBefore
\rowcolor{babyblue!20}{6}
\Body
\toprule
Method &$\mathcal{L}_{\text{rgb}}$&$\mathcal{L}_{\text{3D}}$ &$\mathcal{L}_{\text{temp}}$ &Abs Rel $\downarrow$ 
& Sq Rel $\downarrow$ 
& RMSE $\downarrow$ 
& RMSE log $\downarrow$ 
& $\boldsymbol{\delta}_1$ $\uparrow$ \\
\midrule
{Baseline} & $\times$ & $\times$ &$\times$ 
& {0.100} 
& {2.227} 
& {10.400} 
& {0.255} 
& {0.902} 
\\

\#1 & $\checkmark$ & $\times$ & $\times$ 
& {0.083} 
& {0.825} 
& {7.340} 
& {0.108} 
& {0.955} \\

\#2 & $\checkmark$ & $\checkmark$ & $\times$ 
& {0.074}
& {0.805}
& {7.261}
& {0.098}
& {0.963}
\\

\#3 & $\checkmark$ & $\times$ & $\checkmark$ 
& {0.073}
& {0.700}
& {6.254}
& {0.094}
& {0.952}
\\

Ours & $\checkmark$ & $\checkmark$ & $\checkmark$ 
& \textbf{0.060} 
& \textbf{0.471} 
& \textbf{5.503} 
& \textbf{0.079} 
& \textbf{0.978} 
\\
\bottomrule
\end{NiceTabular}
}
 \vspace{-0.5em}
\end{table*}

\begin{table*}[t]
\centering
\caption{Backbone ablation on SCARED. \textit{Consistency} uses our consistency optimization; VDA uses only $\mathcal{L}_{3D}$ and $\mathcal{L}_{\text{temp}}$ due to RGB rendering limits.} 
\label{tab:ablation_2}
\resizebox{\textwidth}{!}{
\begin{NiceTabular}{ccccccc}
\toprule
Backbone & Consistency & Abs Rel $\downarrow$ 
& Sq Rel $\downarrow$ 
& RMSE $\downarrow$ 
& RMSE log $\downarrow$ 
& $\boldsymbol{\delta}_1$ $\uparrow$ \\
\midrule

\multirow{2}{*}{VDA~\cite{C_2025CVPR_VDA}}& $\times$
& 0.241 
& 7.702 
& 18.673 
& 0.287 
& 0.597  \\

& $\checkmark$& \textbf{0.097} 
& \textbf{1.183} 
& \textbf{8.424} 
& \textbf{0.123}
& \textbf{0.925} \\
\midrule
\multirow{2}{*}{AnySplat~\cite{jiang2025anysplat}} 
& $\times$ 
& {0.100} 
& {2.227} 
& {10.400} 
& {0.255} 
& {0.902} 
\\

&$\checkmark$ 
& \textbf{0.060} 
& \textbf{0.471} 
& \textbf{5.503} 
& \textbf{0.079} 
& \textbf{0.978} 
\\

\bottomrule
\end{NiceTabular}
}
 \vspace{-1em}
\end{table*}
\subsection{Ablation Study}

Ablation results on the SCARED dataset (Table~\ref{tab:ablation_1}) validate the individual and collective contributions of our proposed consistency constraints.
Compared to the AnySplat baseline, the inclusion of photometric ($\mathcal{L}_{\text{rgb}}$) and geometric ($\mathcal{L}_{\text{3D}}$) consistency losses significantly enhances spatial accuracy by anchoring depth predictions to a coherent 3D structure. Furthermore, the temporal consistency loss ($\mathcal{L}_{\text{temp}}$) effectively suppresses flickering and improves stability across video frames. Our full model achieves the best performance across all metrics, demonstrating that these three constraints provide complementary benefits for achieving both high-fidelity geometry and temporal coherence.

We also test the generality of our consistency optimization on different backbones, including VDA~\cite{C_2025CVPR_VDA} and AnySplat~\cite{jiang2025anysplat}. As shown in Table~\ref{tab:ablation_2}, applying our consistency module consistently improves performance across both architectures. For instance, VDA's Abs Rel drops from 0.241 to 0.097 and $\delta$ increases from 0.597 to 0.925, confirming that our method effectively enhances diverse depth estimation backbones.

\section{Conclusion}
In this work, we present a self-supervised 3D consistency optimization framework for endoscopic video depth estimation. Rather than estimating depth independently for each frame, our method exploits the multi-view structure of video to jointly infer depth, camera parameters, and a unified 3D representation via differentiable Gaussian splatting. By incorporating image-level reconstruction, global alignment in the world coordinate system, and temporal consistency, the framework promotes geometrically coherent and temporally stable depth predictions across frames. Results on public benchmarks demonstrate that our framework significantly reduces geometric drift and temporal instability compared to state-of-the-art methods.

\bibliographystyle{splncs04}
\bibliography{string,refs}

@article{jiang2025anysplat,
  title={Anysplat: Feed-forward 3d gaussian splatting from unconstrained views},
  author={Jiang, Lihan and Mao, Yucheng and Xu, Linning and Lu, Tao and Ren, Kerui and Jin, Yichen and Xu, Xudong and Yu, Mulin and Pang, Jiangmiao and Zhao, Feng and others},
  journal={ACM Transactions on Graphics (TOG)},
  volume={44},
  number={6},
  pages={1--16},
  year={2025},
  publisher={ACM New York, NY, USA}
}

@inproceedings{wang2025vggt,
  title={Vggt: Visual geometry grounded transformer},
  author={Wang, Jianyuan and Chen, Minghao and Karaev, Nikita and Vedaldi, Andrea and Rupprecht, Christian and Novotny, David},
  booktitle=CVPR,
  pages={5294--5306},
  year={2025}
}

@inproceedings{C_2025CVPR_VDA,
  title={Video depth anything: Consistent depth estimation for super-long videos},
  author={Chen, Sili and Guo, Hengkai and Zhu, Shengnan and Zhang, Feihu and Huang, Zilong and Feng, Jiashi and Kang, Bingyi},
  booktitle=CVPR,
  pages={22831--22840},
  year={2025}
}

@inproceedings{C_2024MICCAI_endodac,
  title={Endodac: Efficient adapting foundation model for self-supervised depth estimation from any endoscopic camera},
  author={Cui, Beilei and Islam, Mobarakol and Bai, Long and Wang, An and Ren, Hongliang},
  booktitle=MICCAI,
  pages={208--218},
  year={2024},
}

@inproceedings{C_2025MICCAI_endodav,
  title={EndoDAV: Depth Any Video in Endoscopy with Spatiotemporal Accuracy},
  author={Zhou, Zanwei and Yang, Chen and Yang, Piao and Yang, Xiaokang and Shen, Wei},
  booktitle=MICCAI,
  pages={192--201},
  year={2025},
  organization={Springer}
}

@inproceedings{huang2024endo4dgs,
  title     = {Endo-4DGS: Endoscopic Monocular Scene Reconstruction with 4D Gaussian Splatting},
  author={Huang, Yiming and Cui, Beilei and Bai, Long and Guo, Ziqi and Xu, Mengya and Islam, Mobarakol and Ren, Hongliang},
  booktitle = MICCAI,
  pages     = {197--207},
  year      = {2024}
}

@inproceedings{loshchilov2017adamw,
 title         = {Decoupled Weight Decay Regularization},
  author        = {Loshchilov, Ilya and Hutter, Frank},
    booktitle = ICLR,
    year = 2019
}

@misc{oquab2023dinov2,
  title  = {DINOv2: Learning Robust Visual Features without Supervision},
  author = {Oquab, M. and others},
  year   = {2023}
}

@inproceedings{paszke2019pytorch,
  title     = {PyTorch: An Imperative Style, High-Performance Deep Learning Library},
  author    = {Paszke, A. and others},
  booktitle = NIPS,
  volume    = {32},
  year      = {2019},
}

@article{shao2022selfsupervisedendoscopy,
  title   = {Self-Supervised Monocular Depth and Ego-Motion Estimation in Endoscopy: Appearance Flow to the Rescue},
  author={Shao, Shuwei and Pei, Zhongcai and Chen, Weihai and Zhu, Wentao and Wu, Xingming and Sun, Dianmin and Zhang, Baochang},
  journal = {Medical Image Analysis},
  volume  = {77},
  pages   = {102338},
  year    = {2022},
}

@inproceedings{wang2024endogslam,
  title     = {EndoGSLAM: Real-Time Dense Reconstruction and Tracking in Endoscopic Surgeries using Gaussian Splatting},
author={Kailing Wang and Chen Yang and Yuehao Wang and Sikuang Li and Yan Wang and Qi Dou and Xiaokang Yang and Wei Shen},
  booktitle = MICCAI,
  pages     = {219--229},
  year      = {2024}
}

@inproceedings{wei2024enhancedscaleaware,
  title     = {Enhanced Scale-Aware Depth Estimation for Monocular Endoscopic Scenes with Geometric Modeling},
  author    = {Wei, R. and Li, B. and Chen, K. and Ma, Y. and Liu, Y. and Dou, Q.},
  booktitle = MICCAI,
  pages     = {263--273},
  year      = {2024}
}

@inproceedings{yang2023neurallerplane,
  title     = {Neural Lerplane Representations for Fast 4D Reconstruction of Deformable Tissues},
  author    = {Yang, C. and Wang, K. and Wang, Y. and Yang, X. and Shen, W.},
  booktitle = MICCAI,
  pages     = {46--56},
  year      = {2023}
}

@InProceedings{C_2017CVPR_pose-free-1,
author = {Godard, Clement and Mac Aodha, Oisin and Brostow, Gabriel J.},
title = {Unsupervised Monocular Depth Estimation With Left-Right Consistency},
booktitle = CVPR,
year = {2017}
}

@InProceedings{C_2025CVPR_pose-free-2,
    author    = {Kang, Gyeongjin and Yoo, Jisang and Park, Jihyeon and Nam, Seungtae and Im, Hyeonsoo and Shin, Sangheon and Kim, Sangpil and Park, Eunbyung},
    title     = {SelfSplat: Pose-Free and 3D Prior-Free Generalizable 3D Gaussian Splatting},
    booktitle = CVPR,
    year      = {2025},
    pages     = {22012-22022}
}

@inproceedings{C_2025MICCAI_Endo3R,
author="Guo, Jiaxin
and Dong, Wenzhen
and Huang, Tianyu
and Ding, Hao
and Wang, Ziyi
and Kuang, Haomin
and Dou, Qi
and Liu, Yun-Hui",
title="Endo3R: Unified Online Reconstruction from Dynamic Monocular Endoscopic Video",
    booktitle = MICCAI,
    year = 2025,
}

@Article{J_2023ACMToG_3DGS,
      author       = {Kerbl, Bernhard and Kopanas, Georgios and Leimk{\"u}hler, Thomas and Drettakis, George},
      title        = {3D Gaussian Splatting for Real-Time Radiance Field Rendering},
      journal      = {ACM Transactions on Graphics},
      number       = {4},
      volume       = {42},
      year         = {2023},
}

@ARTICLE{J_2021_hamlyn,
  author={Recasens, David and Lamarca, José and Fácil, José M. and Montiel, J. M. M. and Civera, Javier},
  journal={IEEE Robotics and Automation Letters}, 
  title={Endo-Depth-and-Motion: Reconstruction and Tracking in Endoscopic Videos Using Depth Networks and Photometric Constraints}, 
  year={2021},
  volume={6},
  number={4},
  pages={7225-7232},
}

@inproceedings{C_2025MICCAI_3dgs,
  title={Surgicalgs: Dynamic 3d gaussian splatting for accurate robotic-assisted surgical scene reconstruction},
  author={Chen, Jialei and Zhang, Xin and Hoque, Mobarak I and Vasconcelos, Francisco and Stoyanov, Danail and Elson, Daniel S and Huang, Baoru},
  booktitle=MICCAI,
  pages={572--582},
  year={2025},
}

@inproceedings{C_2025MICCAI_foundation,
  title={Endo-fast3r: endoscopic foundation model adaptation for structure from motion},
  author={Sheikh Zeinoddin, Mona and Hoque, Mobarak I and Tandogdu, Zafer and Shaw, Greg L and Clarkson, Matthew J and Mazomenos, Evangelos B and Stoyanov, Danail},
  booktitle=MICCAI,
  pages={117--126},
  year={2025},
}

@inproceedings{C_2025MICCAI_SurgTPGS,
  title={SurgTPGS: Semantic 3D Surgical Scene Understanding with Text Promptable Gaussian Splatting},
  author={Huang, Yiming and Bai, Long and Cui, Beilei and Yuan, Kun and Wang, Guankun and Hoque, Mobarak I and Padoy, Nicolas and Navab, Nassir and Ren, Hongliang},
  booktitle=MICCAI,
  pages={584--594},
  year={2025},
}

@inproceedings{C_2025MICCAI_navigation,
  title={Bridgesplat: Bidirectionally coupled ct and non-rigid gaussian splatting for deformable intraoperative surgical navigation},
  author={Fehrentz, Maximilian and Winkler, Alexander and Heiliger, Thomas and Haouchine, Nazim and Heiliger, Christian and Navab, Nassir},
  booktitle=MICCAI,
  pages={44--53},
  year={2025},
}

@string{CVPR = "Proceedings of the IEEE/CVF Conference on Computer Vision and Pattern Recognition (CVPR)"}

@string{NIPS = "Proceedings of the Annual Conference on Neural Information Processing Systems (NeurIPS)"}

@string{ICLR = "Proceedings of the International Conference on Learning Representations (ICLR)"}

@string{MICCAI = "Proceedings of the International Conference on Medical Image Computing and Computer Assisted Intervention (MICCAI)"}
\end{document}